\documentclass[final]{cvpr}

\usepackage{times}
\usepackage{epsfig}
\usepackage{graphicx}
\usepackage{amsmath}
\usepackage{amssymb}
\usepackage{caption}

\usepackage{subfigure} 
\usepackage{color}
\usepackage[table]{xcolor}
\usepackage{makecell}
\usepackage{numprint}
\usepackage{multirow}
\usepackage{bbm}
\usepackage{float}
\usepackage{comment}

\newcommand{\ignore}[1]{}

\definecolor{green_im}{rgb}{0.0, 0.5, 0.0}

\newcommand{\cvpara}[1]{\vspace{0.05in}\noindent\textbf{#1}}

\definecolor{citecolor}{RGB}{0, 102, 255}

\usepackage[pagebackref=true,breaklinks=true,colorlinks,citecolor=citecolor,bookmarks=false]{hyperref}

\def\cvprPaperID{2757} 
\def\confYear{CVPR 2021}

\begin{document}

\title{Fashionpedia-Ads: Do Your Favorite Advertisements Reveal Your Fashion Taste?}

\author{
  Mengyun Shi$^{1}$\hspace{8pt}
  Claire Cardie$^{1}$\hspace{8pt}
  Serge Belongie$^{2}$\hspace{8pt}\\
$^{1}$Cornell University
\qquad $^{2}$University of Copenhagen
}

\twocolumn[{
\renewcommand\twocolumn[1][]{#1}
\maketitle
\vspace*{-0.55cm}
\includegraphics[width=\linewidth]{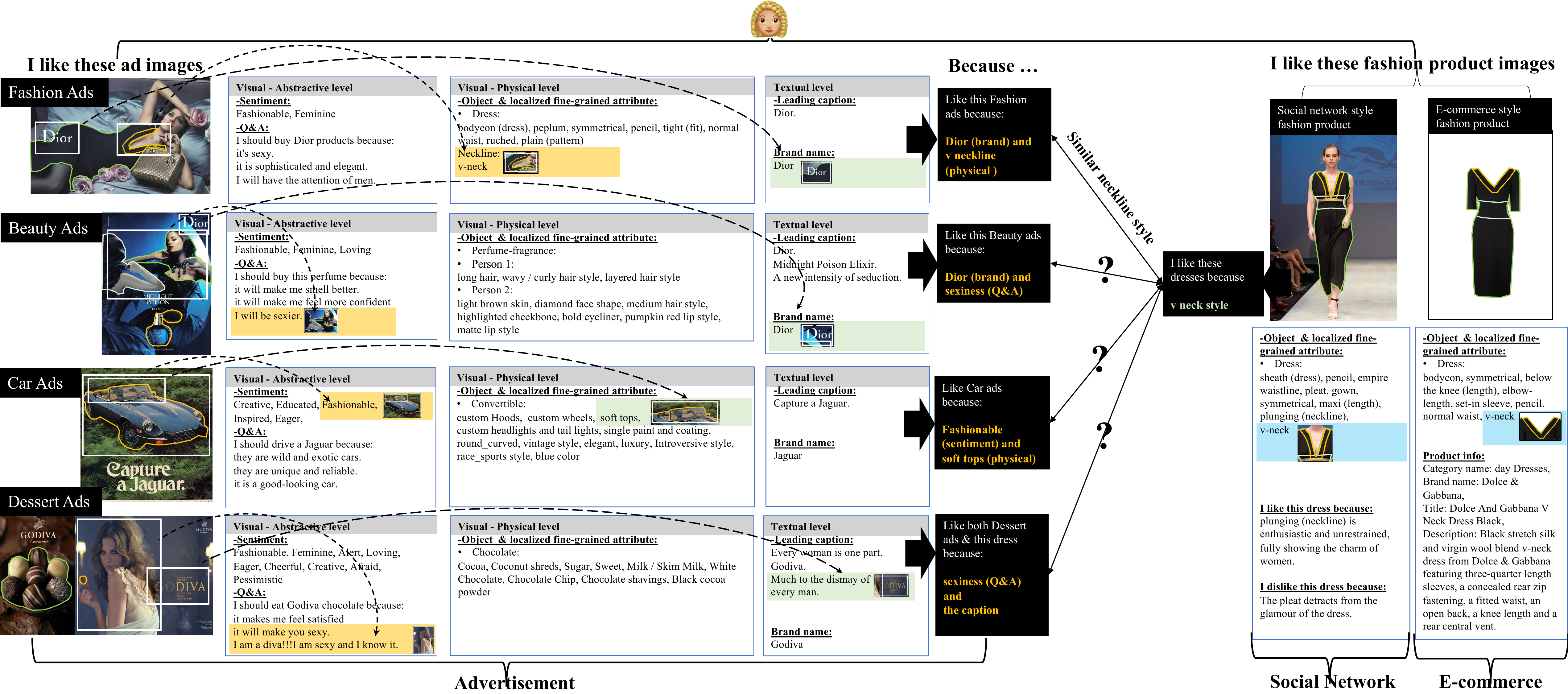}
\captionof{figure}{(a) \emph{Fashion ads V.S. fashion products}: the subject likes both the fashion ads and product images because of the similar V-neckline style of the dresses; (b) \emph{Beauty/car/dessert ads V.S. fashion products}: the subject's preference on these ads can be aroused by multi-perspectives expressed in the ads, such as sentiment, physical attributes or brands. Our Fashionpedia-Ads dataset challenges computer vision systems to not only predict whether
a subject like a fashion product based on given ad images, but also provide a rational explanation why it makes this prediction from multi-perspectives of ads.}
\label{fig:teaser}
\vspace*{0.5cm}
}]

\maketitle


\begin{abstract}
    Consumers are exposed to advertisements across many different domains on the internet, such as fashion, beauty, car, food, and others. On the other hand, fashion represents second highest e-commerce shopping category.
    Does consumers' digital record behavior on various fashion ad images reveal their fashion taste?
    Does ads from other domains infer their fashion taste as well? In this paper, we study the correlation between advertisements and fashion taste.
    Towards this goal, we introduce a new dataset, Fashionpedia-Ads, which asks subjects to provide their preferences on both ad (fashion, beauty, car, and dessert) and fashion product (social network and e-commerce style) images. Furthermore, we exhaustively collect and annotate the emotional, visual and textual information on the ad images from multi-perspectives (abstractive level, physical level, captions, and brands). We open-source Fashionpedia-Ads to enable future studies and encourage more approaches to interpretability research between advertisements and fashion taste. Fashionpedia-Ads can be found at: \footnote{Fashionpedia project page: \href{https://fashionpedia.github.io/home/}{\texttt{fashionpedia.github.io/home/}}}

\end{abstract}

\section{Introduction}
\label{sec: intro}





\begin{figure}
\centering
\includegraphics[width=\columnwidth]{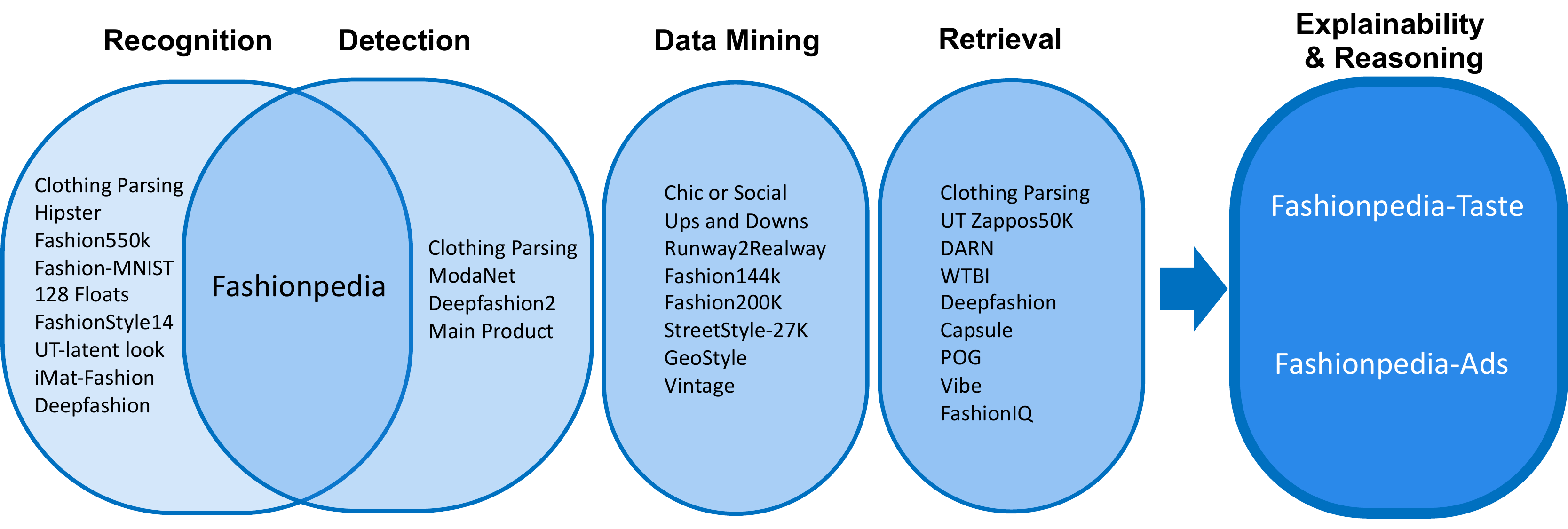}
\caption{Previous fashion datasets focus on recognition~\cite{guo2019imaterialist, inoue_multi-label_2017, jia2020fashionpedia, fleet_hipster_2014, liu2016deepfashion, simo-serra_fashion_2016, takagi_what_2017, xiao_fashion-mnist:_2017, yamaguchi_parsing_2012}, detection~\cite{ge_deepfashion2:_2019, jia2020fashionpedia, yu_multi-modal_2017, zheng_modanet:_2018}, data mining~\cite{han_automatic_2017, he_ups_2016, hsiao_learning_2017, hsiao2021culture,  mall2019geostyle, simo-serra_neuroaesthetics_2015, matzen_streetstyle:_2017, vittayakorn_runway_2015, yamaguchi_chic_2014}, and retrieval~\cite{chen2019pog, hsiao2018creating, hsiao2020vibe, huang_cross-domain_2015, kiapour_where_2015, liu2016deepfashion, wu2021fashion, yu2017semantic}. Fashionpedia-Taste studies fashion taste based on fashion products. Fashionpedia-Ads dataset further investigates fashion taste interpretability and reasoning between advertisements and fashion products.
}
\label{fig:literature_review_ads}
\vspace{-0.4cm}
\end{figure}

It is understandable that there could be some correlation between ads and products for a same domain. For example, a user likes the style of a neckline in a fashion ads and might also like a fashion product that has similar style (Fig.~\ref{fig:teaser}). However, is there any correlation between ads and products from different domains? Specifically, can we interpret a consumer’s product preference from her website browsing logs of various advertising domains? In the context of fashion online shopping, however,
to our knowledge, no study has investigated the correlation between various ads domain and fashion taste on the consumer level, as shown in Fig.~\ref{fig:literature_review_ads}.

In this paper, we introduce a new user taste understanding dataset, Fashionpedia-Ads, which asks subjects to provide their preference on both ad images of various domain (fashion, beauty, car, food) and fashion product images.
Furthermore, unlike fashion product images, ads images usually contains complicated and multiple perspectives of information (emotional, visually, textually...) that cause a consumer like them. For example, for a same ad image (Fig.~\ref{fig:teaser}), a consumer might like it because of the neckline of the dress. However, another consumer might like this ad image because the emotional feeling created in this ad image. 
To fully understand the multi-correlation (both visual and textual) between ads and fashion product images liked by subjects, we exhaustively annotated both ads and fashion images from different perspectives: 1) abstractive level; 2) physical attributes with associated segmentations (localized attributes); 3) caption, and 4) brands on the ads.

The aim of this work is to enable future studies and encourage more exploration to interpretability research between advertisements and fashion taste. The contributions of this work are:
1) we introduce Fashionpedia-Ads, consisting of three datasets (Ads, Social network style and E-commerce style fashion products). We bridge the connection among them through the subjects' preference (like or dislike) on these images and the annotation from multi-perspectives (e.g. abstract \& physical attributes). 2) we formalize a new task that not only requires models to predict whether a subject like or dislike a fashion product image based on given ad images of various domains, but also provide a rationale explaination why it makes this prediction from multi-perspectives.





\section{Dataset Creation Details}
\label{sec:dataset_creation}


\cvpara{3 sub-datasets} Fashionpedia-Ads dataset consisits of 3 sub-datasets. 1) \emph{Advertisement dataset:} this dataset consists of ad images from fashion, beauty, car, and dessert domains that consumers could see on the internet. We use the images from the ads dataset~\cite{hussain2017automatic} for our study;
2) \emph{Social network style fashion product dataset:} this dataset consists of street and runway style fashion product images (with human body), which simulates images that consumers could see on social network websites. We use the images from Fashionpedia dataset~\cite{jia2020fashionpedia} for our study;
3) \emph{E-commerce style fashion product dataset:} this dataset consists of online shopping style fashion product images (without human body), which simulates images that consumers could see on E-commerce websites. We collect the images and associated product information from an online shopping website called nuji.com.

\cvpara{Build the relationship among the 3 sub-datasets through subject preference}
We build the connection among the 3 sub-datasets by asking the same 100 subjects to annotate their preference (like or dislike) for all these 3 sub-datasets, as illustrated in Fig.~\ref{fig:teaser}.

\section{Advertisement Annotation}
\label{supsec:hs_result}

\cvpara{Visual \& Abstractive perspective} We extract the sentiment and question-answer pairs (Action \& Reason) annotation from the ads dataset~\cite{hussain2017automatic} directly. These annotations are collected from MTurk workers and can help us understand subjects' preference from sentiment and emotional level.

\cvpara{Visual \& Physical perspective}
We follow the method proposed by Fashionpedia dataset~\cite{jia2020fashionpedia} and construct taxonomies for the 4 ads domains. By using these taxonomies, we annotate localized objects, sub-objects and fine-grained attributes with associated masks for the ad images of these 4 domains. Why we annotate this? In contrast to the sentimental and emotional perspectives of the ads, a subject's preference on ads can also be impacted by the visual effect of products demonstrated in the ads. With this annotation, we could analyze whether subjects' preference is impacted by the physical perspective of the products shown on the ads.

\cvpara{Textual perspective}
We annotate the leading captions indicated in the ad images with the masks. Because the subjects' preference and emotional feeling can be aroused by the textual information displayed on the ads.

\cvpara{Brand perspective}
We also annotate the brand name indicated in the ad image. The brands and their associated brand culture could also tie to subjects' ads preference. For example, if a subject like fashion products from Dior, she could also like beauty products from Dior, as illustrated in Fig.~\ref{fig:teaser}.


\section{Fashion Taste Annotation}
\label{supsec:data}


To fully understand the contextual information from fashion product images, we annotate the \emph{social network style} and \emph{E-commerce style} fashion product images respectively, as mentioned in Sec.~\ref{sec:dataset_creation}. 1) \emph{The social network style dataset:} we directly use the annotation (task1/2/3/4) created in Fashionpedia-Taste, which asked the 100 subjects to explain their fashion taste in 3 perspectives: a) localized attributes; b) human attention; c) caption, as shown in Fig.~\ref{fig:teaser};
2) \emph{The E-commerce style dataset:} similar to Fashionpedia dataset~\cite{jia2020fashionpedia}, we annotate localized objects, sub-objects and fine-grained attributes with associated masks. Additionally, we annotate the dress length and introversive/extroversive for each image. Each image also contains detailed product information, as shown in Fig.~\ref{fig:teaser}.



\section{Dataset Analysis}
\label{supsec:data_analysis}


\subsection{Visual-Abstractive Attribute Level}

\begin{figure*}[t]
\centering
\subfigure[Fashion Ads - Sentiment.]{
    \includegraphics[scale=0.19]{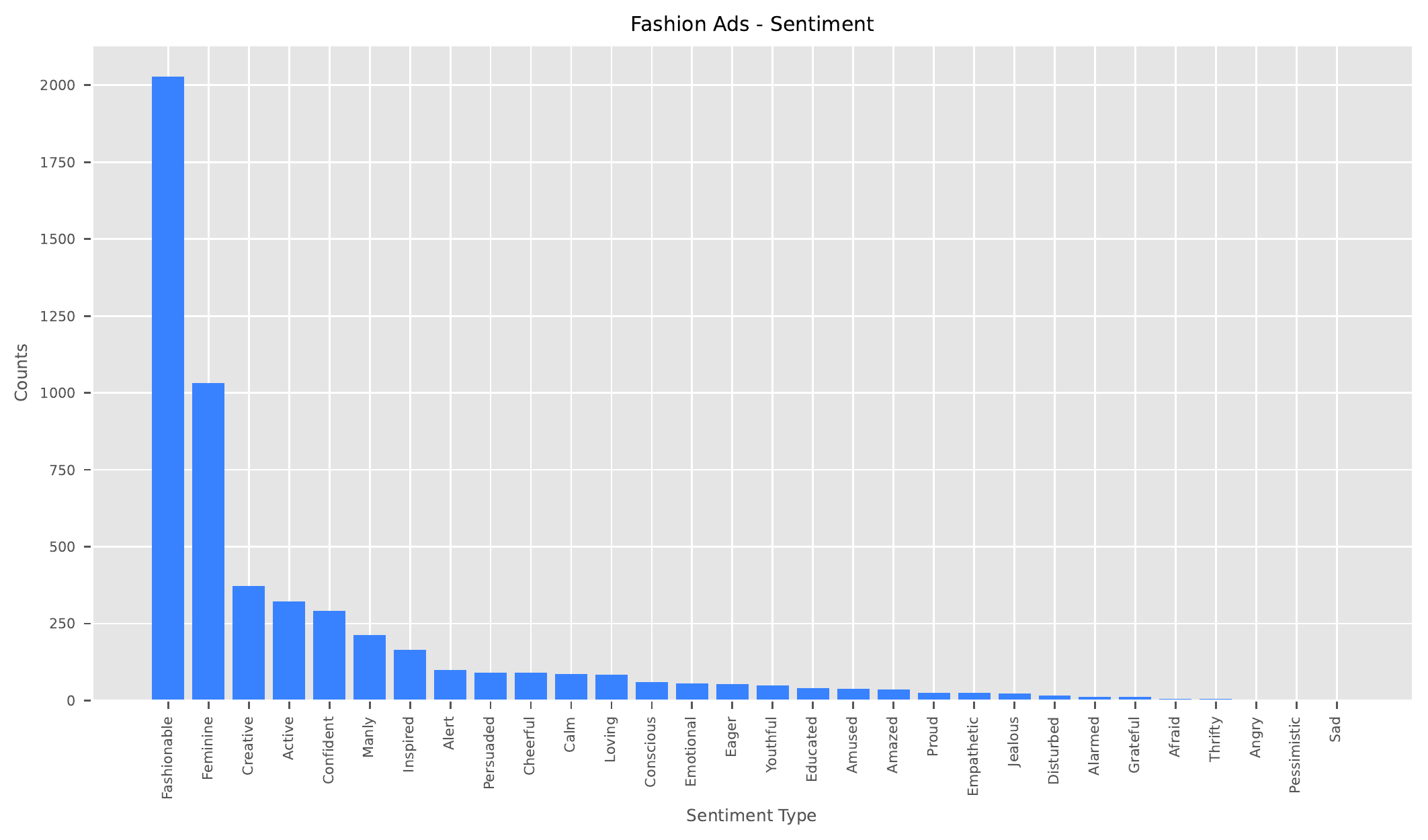}
    \label{fig:Fashion Ads - Sentiment}
}
\hfill
\subfigure[Beauty Ads - Sentiment.]{
    \includegraphics[scale=0.19]{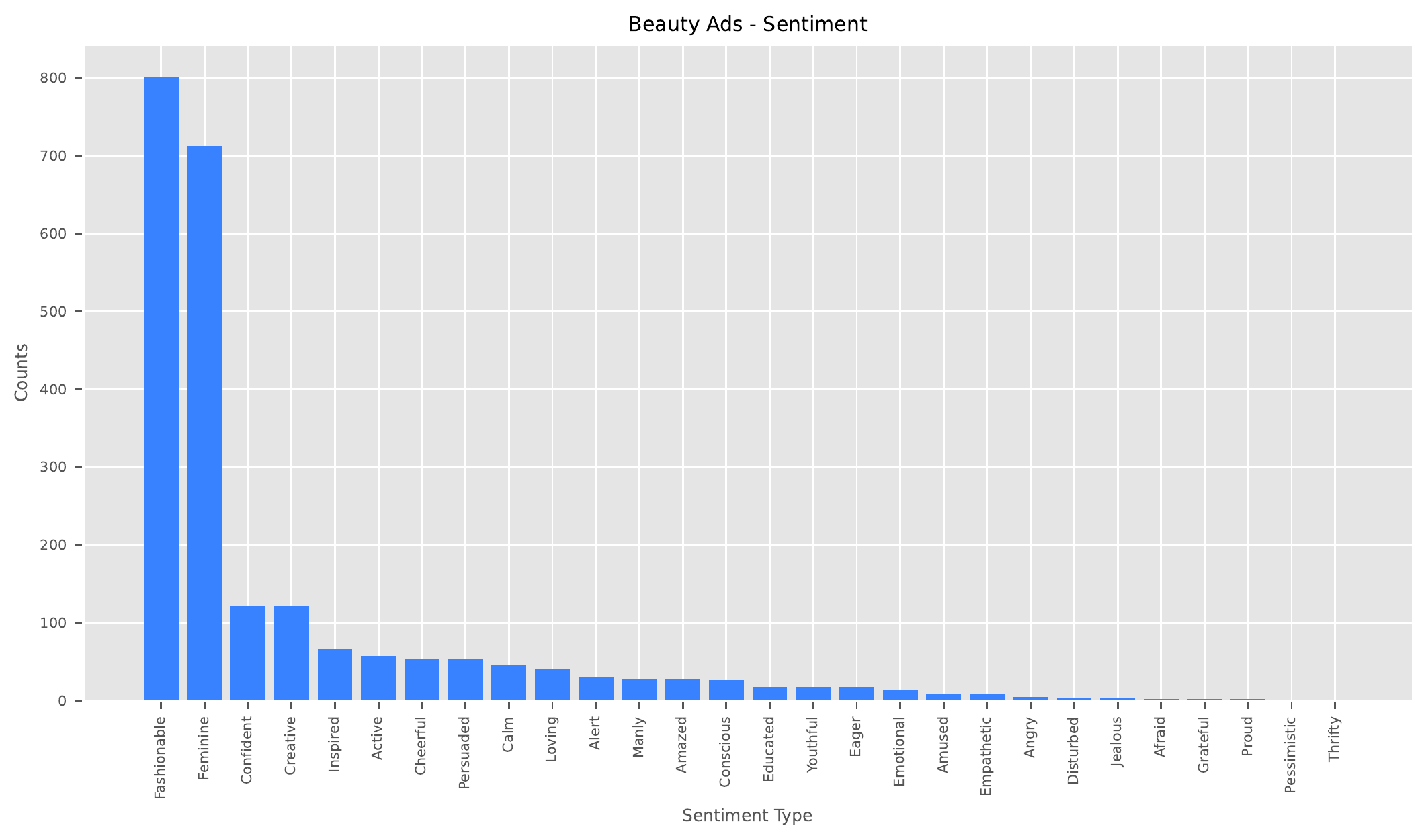}
    \label{fig:Beauty Ads - Sentiment}
}
\hfill
\subfigure[Car Ads - Sentiment.]{
    \includegraphics[scale=0.19]{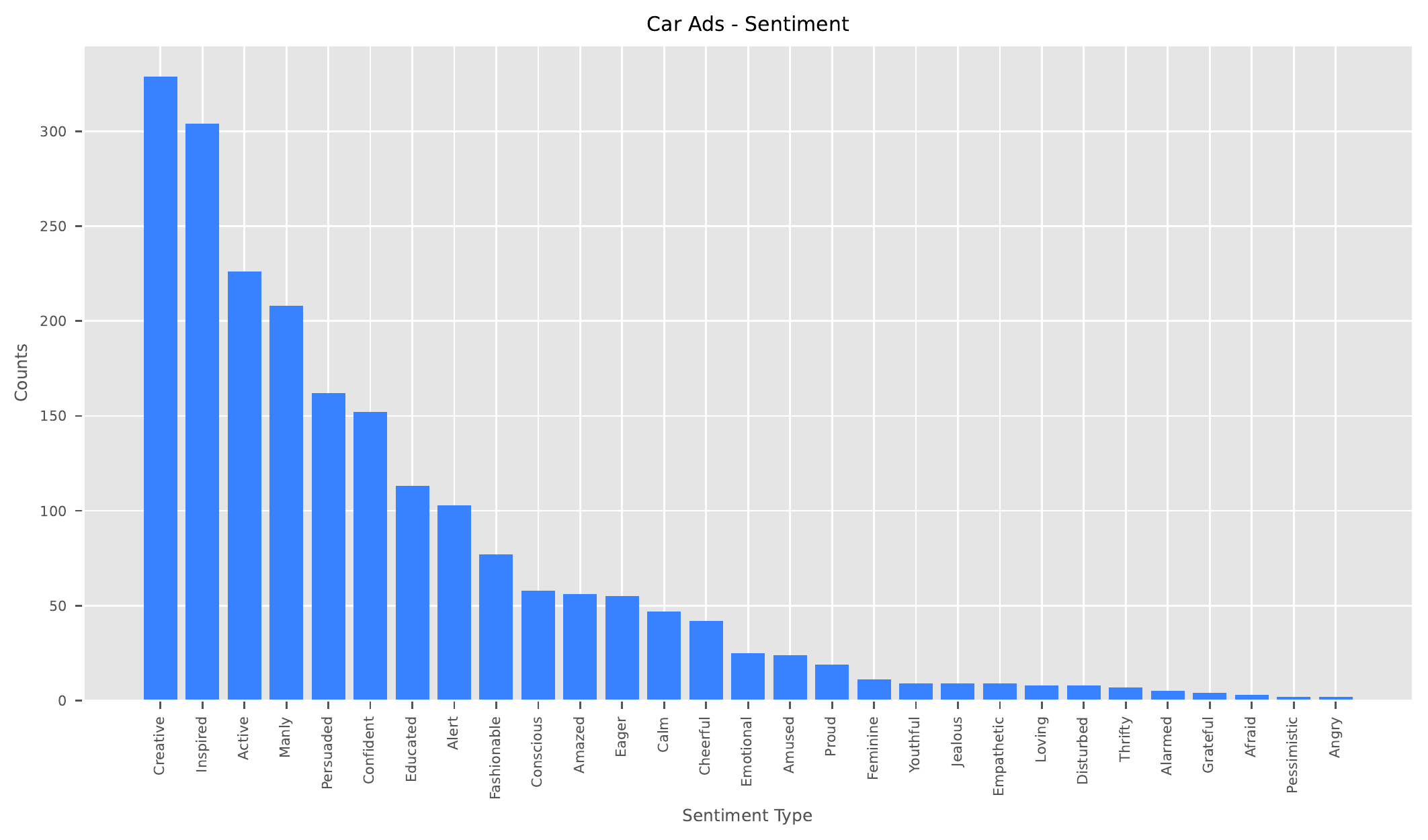}
    \label{fig:Car Ads - Sentiment}
}
\hfill
\subfigure[Dessert Ads - Sentiment.]{
    \includegraphics[scale=0.19]{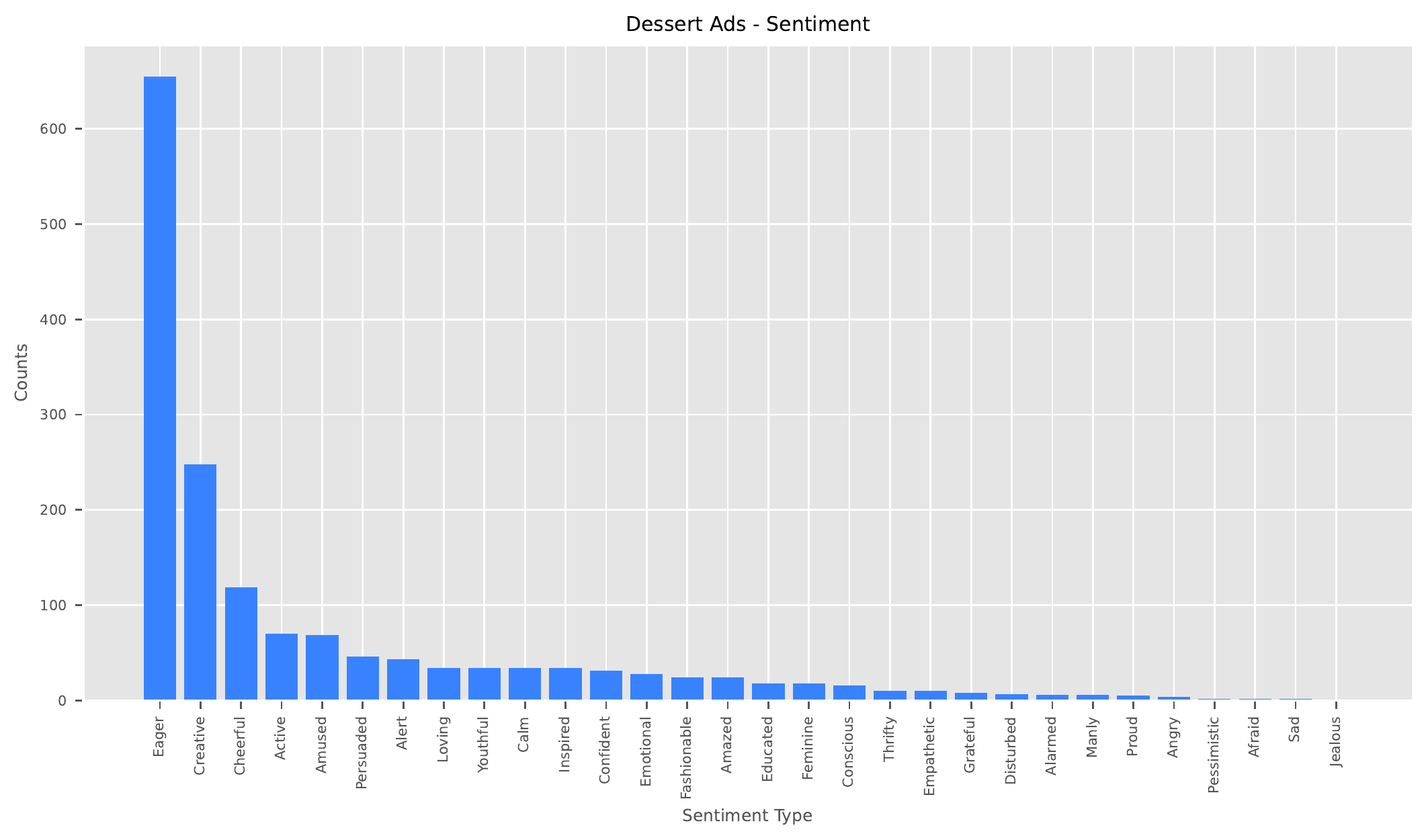}
    \label{fig:Dessert Ads - Sentiment}
}
\caption{
Ads - Sentiment.
}
\vspace{-0.3cm}
\label{fig:Ads - Sentiment}
\end{figure*}

\cvpara{Sentiment}
Fig.~\ref{fig:Ads - Sentiment}
shows the frequency of the sentiment of the ads images for each ads domain. The results show the sentiment from fashion and beauty ads are more correlated compared to car and dessert ads (such as 'fashionable'). Furthermore, fashion and beauty ads are more correlated to 'feminine' sentiment. However, car ads has more tendency of 'manly' sentiment. Dessert ads is invariant of both 'feminine' and 'manly' sentiment. All the 4 ads domains are correlated to 'creative' and 'active' sentiment.

\begin{figure}
\centering
\includegraphics[width=0.85\columnwidth]
{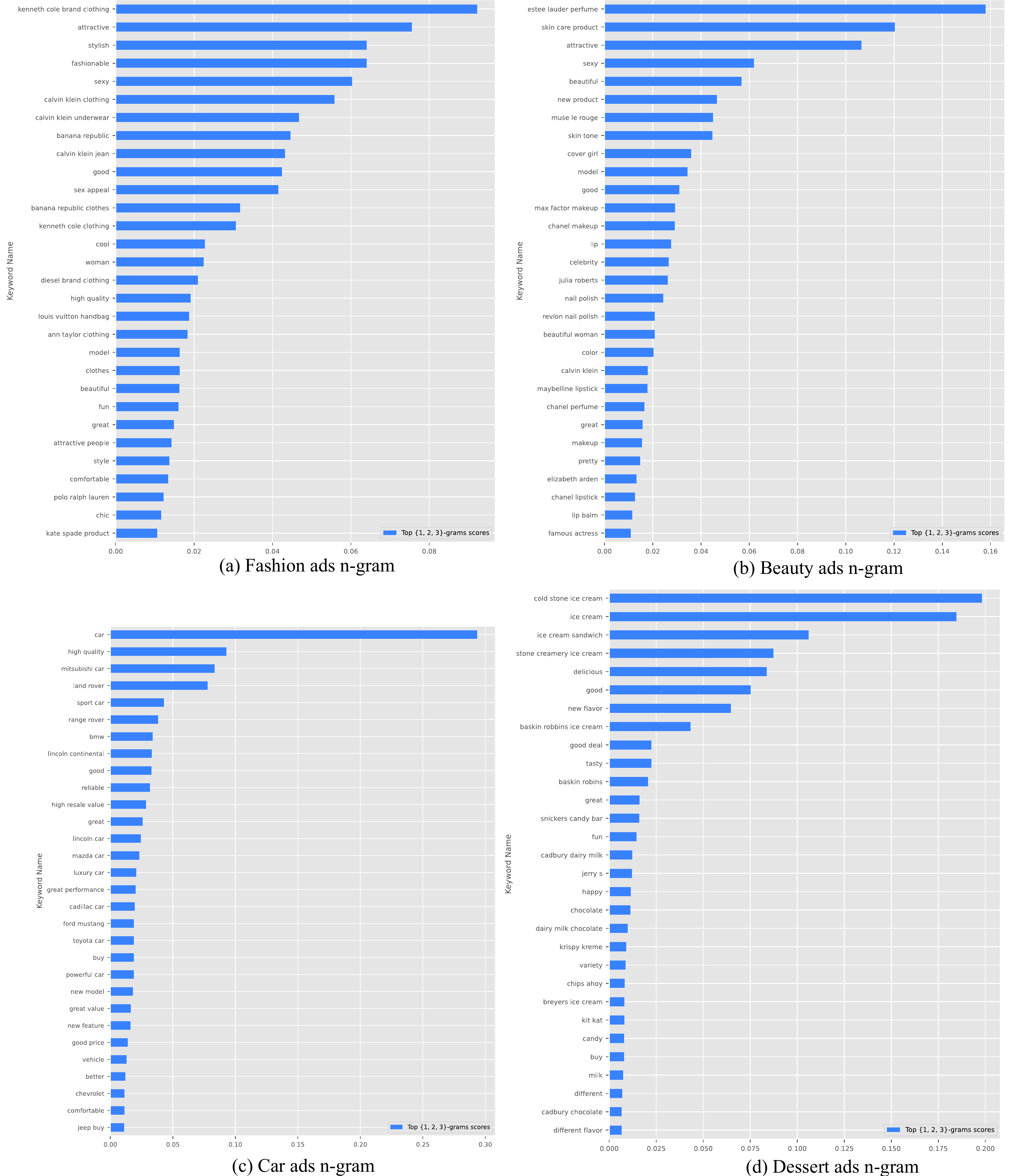}
\caption{QA word count statistics: Number of {1, 2, 3}-grams.}
\label{fig:Ads-QA-ngram}
\vspace{-0.4cm}
\end{figure}

\cvpara{Q\&A word count statistics}
We use SGRank from Textacy ~\cite{textacy2002} to calculate the frequency of words. Fig.~\ref{fig:Ads-QA-ngram} shows the most frequent 1, 2, 3 grams for QA. Similar to the sentiment analysis, fashion and beauty ads share similar emotional feeling. 'Attractive', 'sexy', 'stylish', and 'beautiful' are the most frequent adjectives annotated by MTurk workers. For car ads, the emotional feeling is more related to the functional aspect of cars. The high frequent words include 'high quality', 'reliable', 'great performance', and 'powerful car'. For dessert ads, the emotional words are more connected to the dessert flavors, such as 'delicious', 'new flavor', and 'tasty'.


\cvpara{Q\&A linguistic statistics}
We use part-of-speech (POS) tagging from Spacy~\cite{spacy2002} to tag nouns and adjectives in QA.
Table~\ref{tab:QA Linguistic statistics - Noun} shows the number of most frequent unique nouns by POS. We find the most frequent common nouns is more associated with high level description of a product line indicated in ads images, such as clothes, jeans, perfume, lipstick, ice cream, and cookies. In contrast, the most frequent proper nouns is more related to the brands mentioned in the ads images, such as Gucci, Chanel, Audi, and Häagen-Dazs. This shows the linguistic diversity of our dataset.



\begin{table}[t]
\small
\begin{center}
\resizebox{0.97\columnwidth}{!}{%
\begin{tabular}{ l l l l}
\Xhline{1.0pt}\noalign{\smallskip}
\textbf{Domain}
& \textbf{Word - Noun}   \\ 
\Xhline{1.0pt}\noalign{\smallskip}
Fashion-C & clothes, model, people, style, clothing, classic, jeans, brand, products\\
Fashion-P & kenneth cole, shop, gucci, chanel, gap, kate, ralph lauren, dior\\
\hline
Beauty-C & skin, lips, lashes, model, eyes, colors, celebrity, perfume, lipstick\\
Beauty-P & chanel, julia, rihanna, emma, hello kitty, maybelline, revlon, dior\\
\hline
Car-C & quality, fun, luxury, performance, features, weather, oldsmobile, jeep\\
Car-P & audi, honda, bmw, porsche, subaru, jeep, suv, egypt, toyota, ford\\
\hline
Dessert-C & ice cream, flavors, milk, candy, hunger, variety, snack, cookies\\
Dessert-P & cinnamon, strawberry, almond, häagen dazs, hershey, snickers\\
\Xhline{1.0pt}\noalign{\smallskip}
\end{tabular}
}
\caption{Q\&A Linguistic statistics: unique nouns by POS. '-C ' means common noun. '-P' means proper noun.}
\label{tab:QA Linguistic statistics - Noun}
\vspace{-0.5cm}
\end{center}
\end{table}

\subsection{Visual-Physical Attribute Level}

\begin{figure}
\centering
\includegraphics[width=0.85\columnwidth]
{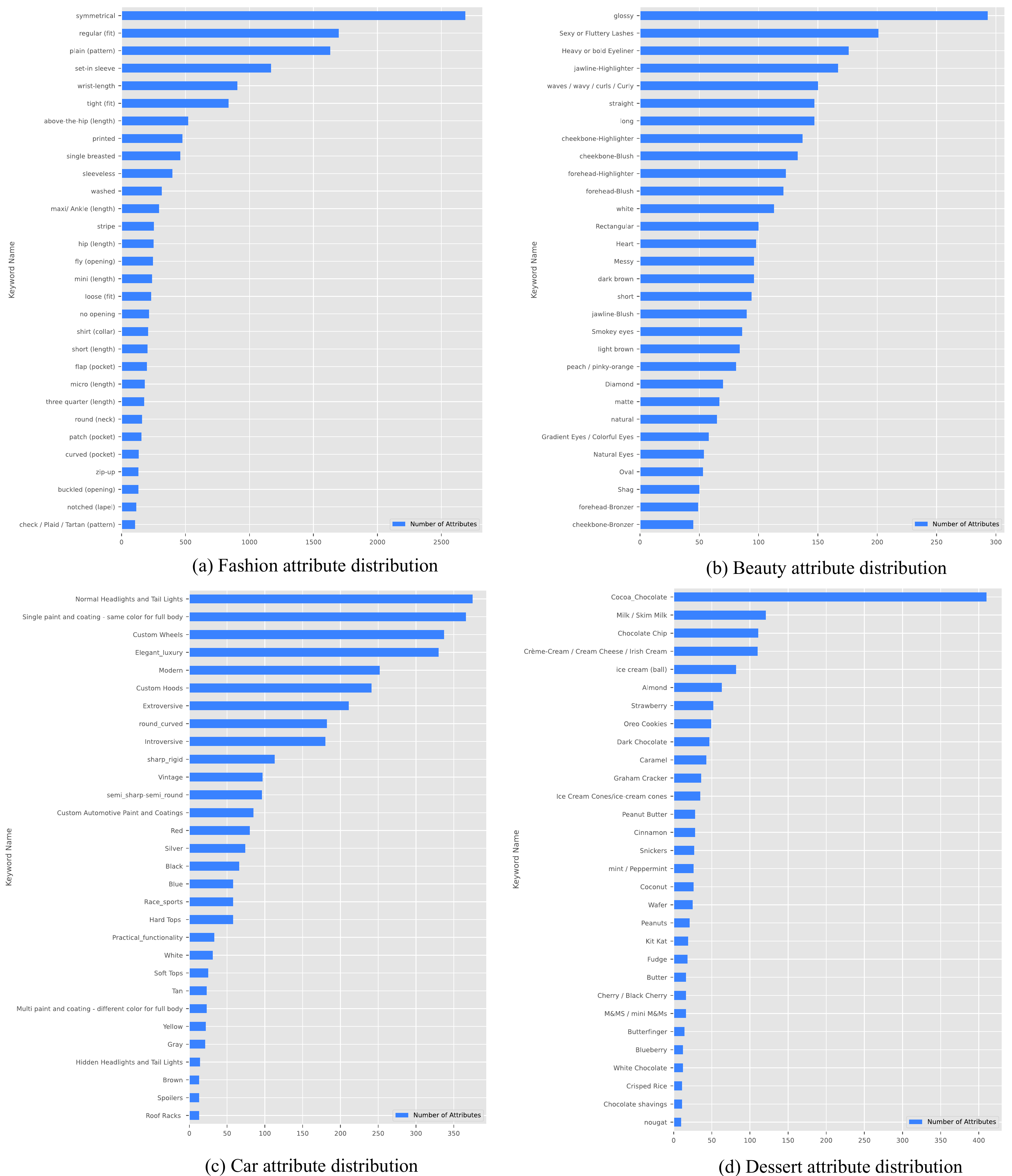}
\caption{Attribute distribution.}
\label{fig:Attribute_label_distribution_index_Name}
\vspace{-0.4cm}
\end{figure}

\cvpara{Localized Attribute distribution}
Fig.~\ref{fig:Attribute_label_distribution_index_Name} shows the distribution of attributes annotated in the 4 ads domains. For fashion ads, 'symmetrical' has highest frequency because most of the garments have balanced silhouette. For beauty ads, 'glossy' has high frequency because the high frequency of lipstick objects in the beauty ads. The similar observation is found for 'cocoa chocolate' in dessert ads. For car ads, 'elegant \& luxury' style has high frequency, which could correlate to the fashion ads and products with 'elegant \& luxury' feeling.


\subsection{Textual information}

\begin{figure}
\centering
\includegraphics[width=0.85\columnwidth]
{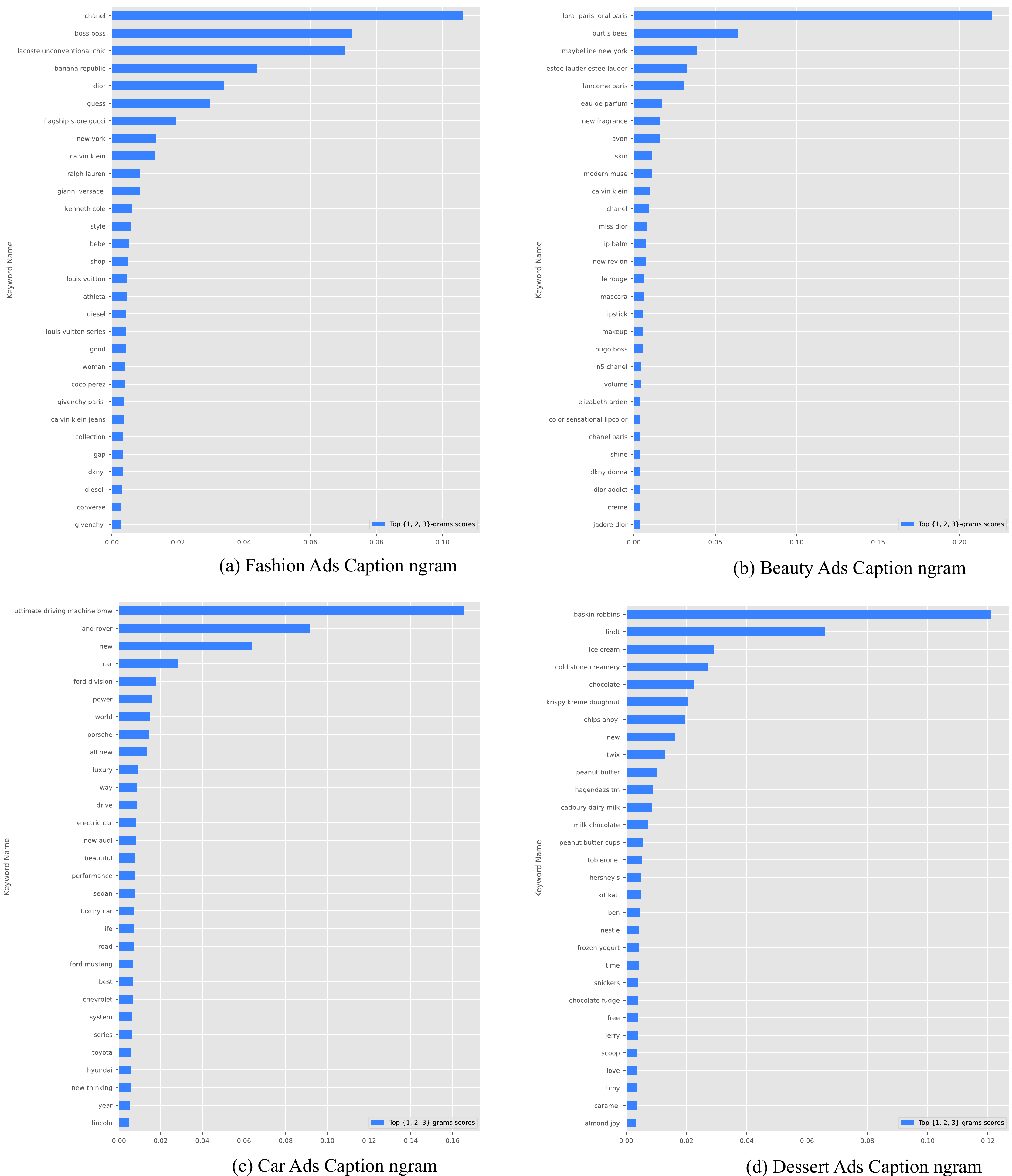}
\caption{Caption word count statistics: Number of 1, 2, 3-grams.}
\label{fig:Ads-Caption-ngram}
\vspace{-0.4cm}
\end{figure}

\cvpara{Ads caption - Word count statistics }
For Fashion ads, Fig.~\ref{fig:Ads-QA-ngram} shows Kenneth cole, Calvin Klein, and Banana republic are most mentioned brands from emotional perspective. However, Chanel, Boss and Lacoste are most frequent brands according to the caption written on the ads images (Fig.~\ref{fig:Ads-Caption-ngram}). This indicates Kenneth cole, Calvin Klein, and Banana republic ads might contain some more provocative visual or textual information that arouses the MTurk workers' sentiment. The similar observation is also found in beauty, car, and dessert ads.

\cvpara{Ads caption - Linguistic statistics}
Table~\ref{tab:Caption Linguistic statistics - Noun} shows number of most frequent nouns from ads caption. Compared to QA (Table~\ref{tab:QA Linguistic statistics - Noun}) annotated from emotional perspective, the caption written on the ad images is more concentrated to brand related keywords and less diverse. This is expected since the signal from QA contains more expression of the viewers' emotional feeling and the caption written on the ads images is more focused on describing the brand itself or products that are advertised in the images.


\begin{table}[t]
\small
\begin{center}
\resizebox{0.97\columnwidth}{!}{%
\begin{tabular}{ l l l l}
\Xhline{1.0pt}\noalign{\smallskip}
\textbf{Domain}
& \textbf{Word - Noun}   \\ 
\Xhline{1.0pt}\noalign{\smallskip}
Fashion-C & dior, boss, gap, hanes, converse, living, lacoste, vera wang, mycalvins\\
Fashion-P & guess, chanel, new, diesel, ralph lauren, calvin klein, banana republic\\
\hline
Beauty-C & volume, paris, skin, boss, fragance, shiseido, lips, serum, hair, revlon\\
Beauty-P & revlon, maybelline, nivea, chanel, dior, lancome, avon, burts bees\\
\hline
Car-C & drive, road, life, oldsmobile, chevrolet, luxury, performance, wheel\\
Car-P & bmw, audi, toyota, volvo, ford, motors, honda, kia, nissan, machine\\
\hline
Dessert-C & toblerone, snickers, doughnuts, cream, lindt, candy, dove, cookies\\
Dessert-P & häagen-dazs, oreo, baskin robbins, nestle, lindt, hershey, cadbury\\
\Xhline{1.0pt}\noalign{\smallskip}
\end{tabular}
}
\caption{Caption linguistic statistics: unique nouns by POS. '-C ' means common noun. '-P' means proper noun.}
\label{tab:Caption Linguistic statistics - Noun}
\vspace{-0.5cm}
\end{center}
\end{table}

\subsection{Brand name}
\begin{figure}
\centering
\includegraphics[width=0.95\columnwidth]
{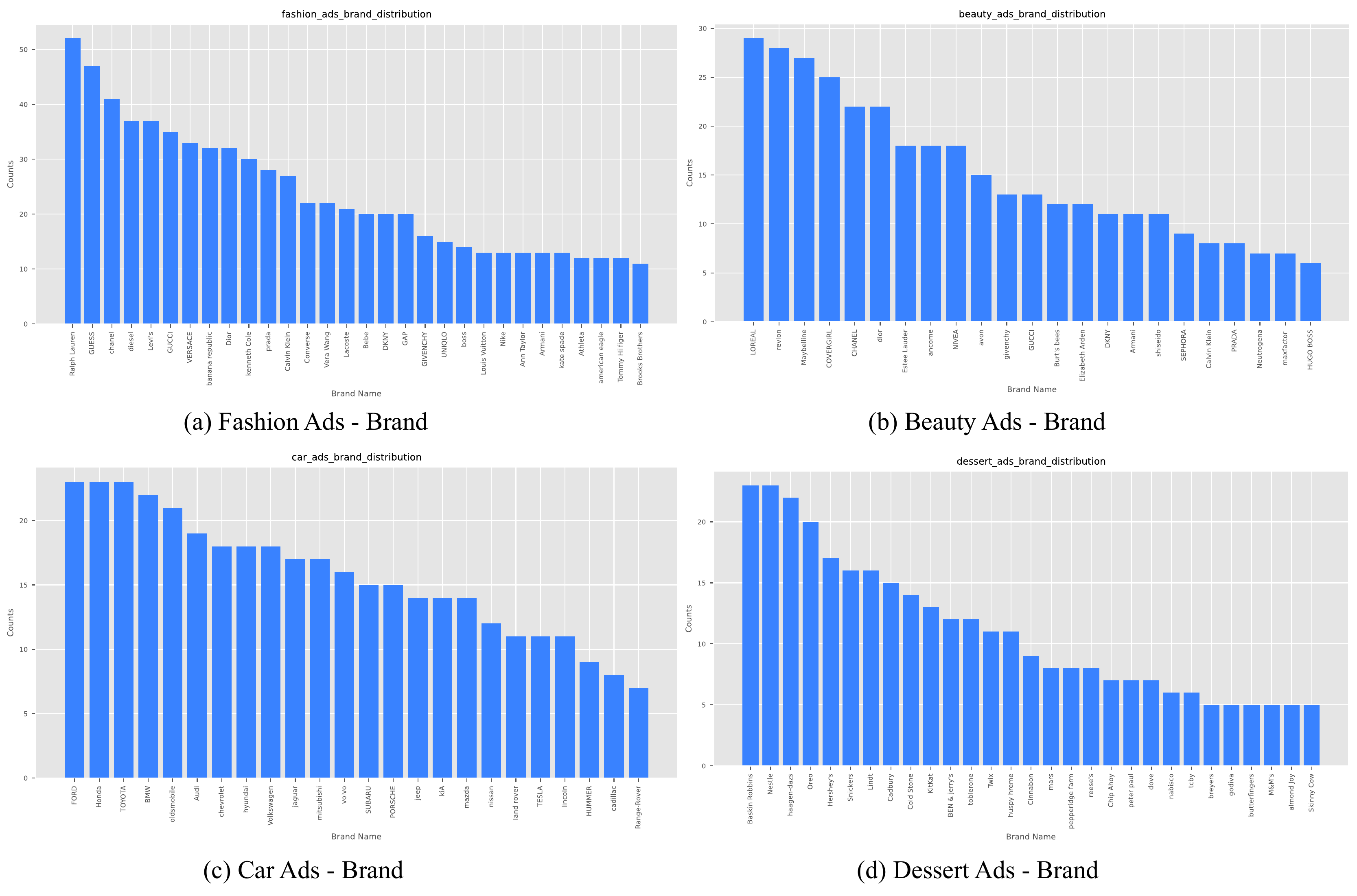}
\caption{Ads - Brand distribution.}
\label{fig:Ads - Brand}
\vspace{-0.4cm}
\end{figure}

Fig.~\ref{fig:Ads - Brand} shows the distribution of brand for all 4 ads domains. We found the brands with highest frequency mentioned in the ads images do not necessary always correspond to the highest brand keywords mentioned in QA (Fig.~\ref{fig:Ads-QA-ngram}) and ads caption (Fig.~\ref{fig:Ads-Caption-ngram}). Fig.~\ref{fig:Ads-QA-ngram} shows Kenneth cole, Calvin Klein, and Banana republic as most mentioned brands from emotional perspective. On the other hand, Chanel, Boss and Lacoste are the most frequent brands according to the caption written on the ads images (Fig.~\ref{fig:Ads-Caption-ngram}). However, the brand distribution ((Fig.~\ref{fig:Ads - Brand})) shows Ralph Lauren has the highest number of images. This indicates some brands might contain some more provocative visual or textual information that arouses the viewers' feeling even they have relatively lower number of images in the dataset. This needs to be further explored in this study. The similar observation is also found in beauty, car, and dessert ads.


\section{Conclusion}

In this work, we studied the problem of modeling human fashion taste in advertisements. We exhaustively collect and annotate the emotional, visual and textual information on the ad images from multi-perspectives (abstractive level, physical level, captions, and brands). We therefore hope that Fashionpedia-Ads will facilitate future research of interpretability between advertisements and fashion taste.



{\small
\bibliographystyle{ieee_fullname}
\bibliography{egbib}
}

\end{document}